\documentclass[journal,transmag]{IEEEtran}
\usepackage{algorithm} 
\usepackage{algorithmic} 
\usepackage{amsfonts}
\usepackage{amssymb}
\usepackage{amsmath}
\usepackage{xcolor}

\usepackage{bbding}

\usepackage[switch,pagewise]{lineno}
\usepackage{colortbl,booktabs}
\usepackage{threeparttable}

\ifCLASSINFOpdf
    \usepackage[pdftex]{graphicx}
    \graphicspath{{../pdf/}{../jpeg/}}
    \DeclareGraphicsExtensions{.pdf,.jpeg,.png}
\else
    \usepackage[dvips]{graphicx}
    \graphicspath{{../eps/}}
\fi
\begin{document}
%
\title{Lesion detection and Grading of Diabetic Retinopathy via Two-stages Deep Convolutional Neural Networks}


\author{\IEEEauthorblockN{Yehui Yang\IEEEauthorrefmark{1,2},
Tao Li\IEEEauthorrefmark{2},
Wensi Li\IEEEauthorrefmark{3},
Haishan Wu\IEEEauthorrefmark{1},
Wei Fan\IEEEauthorrefmark{1} and
Wensheng Zhang\IEEEauthorrefmark{2}}

\IEEEauthorblockA{\IEEEauthorrefmark{1}Big Data Lab, Baidu Research, Baidu, Beijing, 100085, China}
\IEEEauthorblockA{\IEEEauthorrefmark{2}Institute of Automation, Chinese Academy of Sciences, Beijing, 100190, China}
\IEEEauthorblockA{\IEEEauthorrefmark{3}The Beijing Moslem's Hospital, Beijing, 100054, China}
\thanks{Corresponding author:Yehui Yang (email: yangyehuisw@126.com).}}

%



\IEEEtitleabstractindextext{%
\begin{abstract}

 We propose an automatic diabetic retinopathy (DR) analysis algorithm based on two-stages deep convolutional neural networks (DCNN). Compared to existing DCNN-based DR detection methods, the proposed algorithm have the following advantages: (1) Our method can point out the location and type of lesions in the fundus images, as well as giving the severity grades of DR. Moreover, since retina lesions and DR severity appear with different scales in fundus images, the integration of both local and global networks learn more complete and specific features for DR analysis. (2) By introducing imbalanced weighting map, more attentions will be given to lesion patches for DR grading, which significantly improve the performance of the proposed algorithm. In this study, we label $12,206$ lesion patches and re-annotate the DR grades of $23,595$ fundus images from Kaggle competition dataset. Under the guidance of clinical ophthalmologists, the experimental results show that our local lesion detection net achieve comparable performance with trained human observers, and the proposed imbalanced weighted scheme also be proved to significantly improve the capability of our DCNN-based DR grading algorithm.

\end{abstract}

}

\maketitle

\IEEEdisplaynontitleabstractindextext

%
\IEEEpeerreviewmaketitle

\section{Introduction}

Diabetics is an universal chronic disease around some developed countries and developing countries including China and India \cite{Shaw2010,Pratt2016,Haloi2016}. The individuals with diabetic have high probabilistic for having diabetic retinopathy (DR) which is one of the most major cause of irreversible blindness \cite{googleJAMA,Kocur2002}. Therefore, the quickly and automatically detecting of DR is critical and urgent to reduce burdens of ophthalmologist, as well as providing timely morbidity analysis for massive patients.

According to the {\em International Clinical Diabetic Retinopathy Disease Severity Scale} \cite{googleJAMA,DRstages}, the severity of DR can be graded into five stages: normal, mild, moderate, severe and proliferative. The first four stages can also be classified as non-proliferative DR (NPDR) or pre-proliferative DR, and NPDR may turn to proliferative DR (PDR) with high risk if without effective treatment. The early signs of DR are some lesions such as microaneurysm (MA), hemorrhages, exudate etc. Therefore, lesion detection is a less trivial step for the analysis of DR. There are plenty of literatures focus on detecting lesions in retina. Haloi {\em et al.} \cite{Haloi2015} achieve promising performance in exudates and cotton wool spots detection. Later, Haloi \cite{Haloi2016} try to find MAs in color fundus images via deep neural networks. van Grinsven {\em et al.} \cite{TMI2016} propose a selective sampling method for fast hemorrhage detection. Additionally, Srivastava {\em et al.} \cite{Srivastava2017} achieve robust results in finding MA and hemorrhages based on multiple kernel learning method.


However, the aforementioned algorithms do not attach the DR severity grades of the input fundus images, which is vital for the treatment of DR patients. Recently, Seoud {\em et al.} \cite{seoudMICCAI2015} propose an automatic DR grading algorithm based on random forests \cite{randomForest}. By leveraging deep learning techniques \cite{Gu2016}, Gulshan {\em et al.} \cite{googleJAMA} take efforts to classify the fundus images into normal and referable DR (moderate and worse DR) with the annotations of 54 Unite States licensed ophthalmologists on over 128 thousands fundus images. Similarly, Sankar {\em et al.} \cite{Sankar2016} using DCNN to grade DR into normal, mild DR and several DR. Pratt  {\em et al.} \cite{Pratt2016} predict the severity of DR according to the five-stages standard of {\em International Clinical Diabetic Retinopathy Disease Severity Scale}\cite{googleJAMA,DRstages}. Even though these DR grading algorithms seem to have achieved promising performance, they still have the following problems:

(1) The aforementioned DCNN-based DR grading methods can only output the DR grade but cannot indicate the location and type of the existing lesions in the fundus images. However, the detailed information about the lesions may be more significant than a black box for clinicians in treatment.

(2) The above end-to-end DCNN \footnote{End-to-end DCNN grading means that directly feed the input images into DCNN, then output the DR grades of the images.} may not suitable to learn features for DR grading. Compared to the size of the input image, some tiny lesions (eg., MAs and some small hemorrhages) are such unconspicuous that they are prone to be overwhelmed by the other parts of input image via end-to-end DCNN. However, these lesions are critical for DR grading according to the international standard \cite{DRstages}.

To address the above issues, we proposed two-stages DCNN for both lesion detection and DR grading. Accordingly, our method composed of two parts: local network to extract local features for lesion detection and global network to exploit image features in holistic level for DR grading.

Instead of end-to-end DR grading, we construct a weighted lesion map to differentiate the contribution of different parts in image. The proposed weighted lesion map gives imbalanced attentions on different locations of the fundus image in terms of the lesion information, i.e., the patches with more severe lesions will attract more attention to train the global grading net. Such imbalanced weighted scheme significantly improve the capability of the DR grading algorithm (See Section \ref{sec.exp_dr_grading}).

Compared to the existing DCNN-based DR detection algorithms, the proposed algorithm has the following advantages and contributions:

\begin{itemize}
\item[(1)] We propose a two-stages DCNN-based algorithm which can not only detect the lesions in fundus images but also grade the severity of DR. The two-stages DCNNs learn more complete deep features of fundus images for DR analysis in both global and local scale.
\item[(2)] We introduce imbalanced attention on input images by weighted lesion map to improve the performance of DR grading network. To the best of our knowledge, this is the first DNN-based work resorting imbalanced attention to learn underlying features in fundus images for DR grading.
\end{itemize}



\section{Related Work}

Our work is closely related to two topics: DCNN and DCNN-based DR detection. Extensive work has been done on these issues over the past years, and good reviews can be found in \cite{Gu2016,Wong2016}. In this section, we only discuss some most relevant methods to our work.

DCNN is a powerful deep learning architecture inspired by visual mechanisms of animal \cite{Gu2016}. The basic components of DCNN are convolution layer, subsampling layer, fully connected layer, and between two adjacent layer, there may exist an activation functions. A loss function for specific tasks may assigned at the end of network. Additionally, some useful tricks often be utilized to improve the performance of DCNN including Dropout \cite{Dropout}, batch normalization (BN) \cite{BN}, local response normalization \cite{Alexnet}. Due to the reported performance of DCNN, many researchers have taken efforts on this topic and give rise to some representative networks: LeNet-5 \cite{LeNet}, AlexNet \cite{Alexnet}, ZFNet \cite{ZFNet}, VGGNet \cite{VGGNet}, GoogleNet \cite{Googlenet,Szegedy2015} and ResNet \cite{ResNet}. Nowadays, DCNN has been successfully applied in almost all the computer vision tasks  and achieved state-of-the-art performance, such as image classification \cite{ResNet}, object detection \cite{RCNN,FastRCNN,FasterRCNN}, image segmentation \cite{FCN}, action recognition \cite{Oquab2014}, visual saliency detection \cite{Wang2015}.

Recently, automatic DR detection via fundus images has attracted more and more attentions by both researchers in clinical department and computer sciences. Researchers from Google Research and some hospitals used Inception-v3 \cite{Szegedy2015} to detect referable DR and macular edema. In a recent Kaggle competition \footnote{Kaggle DR competition url: https://www.kaggle.com/c/diabetic-retinopathy-detection}, all the top-5 teams had applied deep learning based algorithms to grade the severity of DR. The wining entry had claimed to achieve the comparable performance with ophthalmologist \cite{Szegedy2015,kaggleHuman}. Later, Alban and Gilligan \cite{stanford2016} took the comparison between AlexNet and GoogleNet to end-to-end grade on Kaggle DR dataset, and some other researchers constructed they own DCNN architectures for end-to-end DR grading \cite{Pratt2016,Sankar2016}.

However, the above DCNN DR detection methods are all black box for DR grading, i.e, they do not provide the information of the lesions in the DR screening. To analyze DR in lesion scale, van Grinsven {\em et al.} \cite{TMI2016} found the existence of hemorrhages within the color fundus images. Haloi \cite{Haloi2016} trained a DCNN to find the MAs. Lim {\em et al.} \cite{LimAAAI2014} proposed a DCNN-based transformed representations for lesion detection in retinal images. Additionally, Antal and Hajdu \cite{Antal2014} addressed both MA detection and DR grading via an ensemble-based algorithm.

Inspired by the superior performance of DCNN in computer vision and DR detection, in this study, we propose a two-stages DCNN for DR analyzing. The proposed local and global networks extract the deep features of fundus images in both local and global scale, and the lesion information exploited by the local net improve the correctness of global DR grading net.


\section{Methods}

\begin{figure*}[!t]
\centering
\includegraphics[width=6.5in]{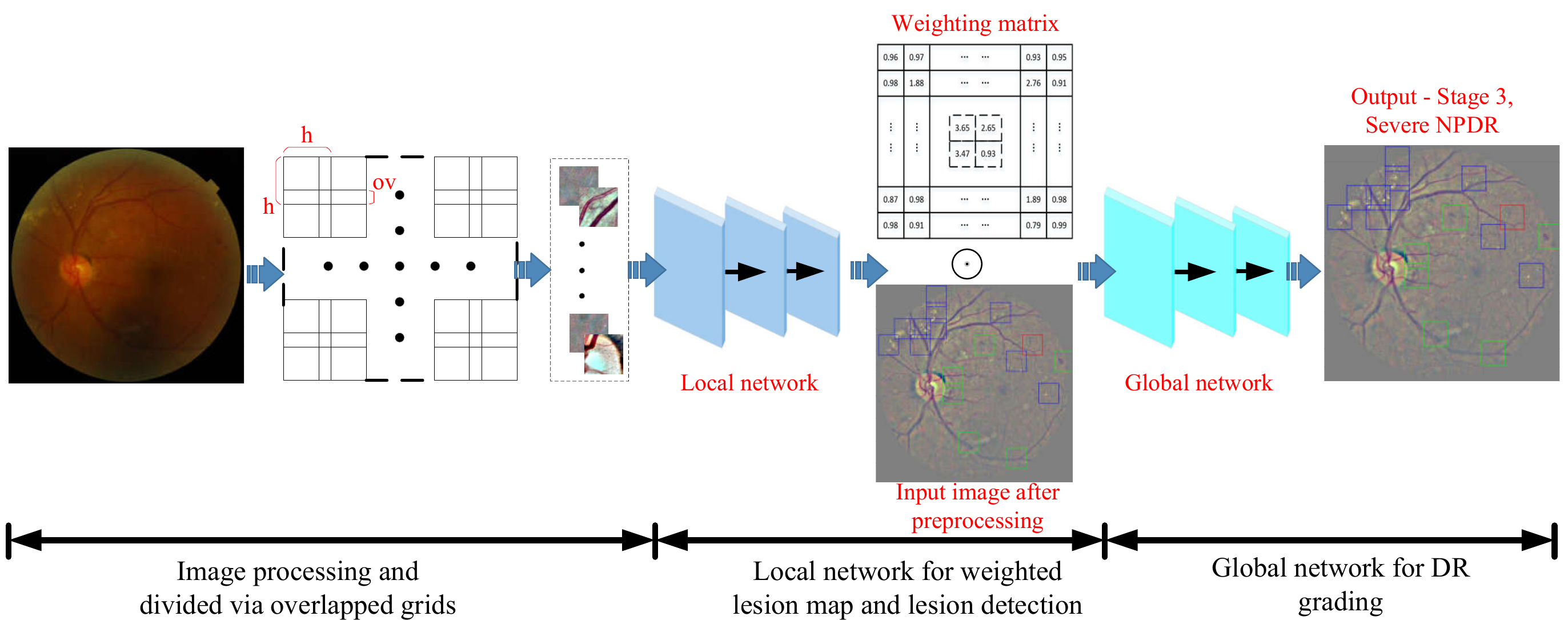}
\caption{Main workflow of the proposed algorithm}
\label{fig.mainWorkflow}       
\end{figure*}

In this section, we present the details of the proposed two-stages DCNN for lesion detection and DR grading. The main workflow of our algorithm is illustrated in Fig. \ref{fig.mainWorkflow}.
First, the input fundus photographs image will be preprocessed and divided into patches. Then, the weighted lesion map is generated based on the input image and output of local net. Third, the proposed global network is introduced for grading the DR severity of input fundus image.

\subsection{Image Division Via Overlap Grids}

Image division is a critical and essential step for providing lesion candidates and generating the lesion map. In most detection tasks, detection candidates are provided by selective search \cite{selectiveSearch}, which ignore the relative location among the candidates in single image. However, the relative location among patches is less trivial in the construction of weight lesion map (see Section \ref{subsec.lessionMap}).

\begin{figure}[!t]
\centering
\includegraphics[width=3in]{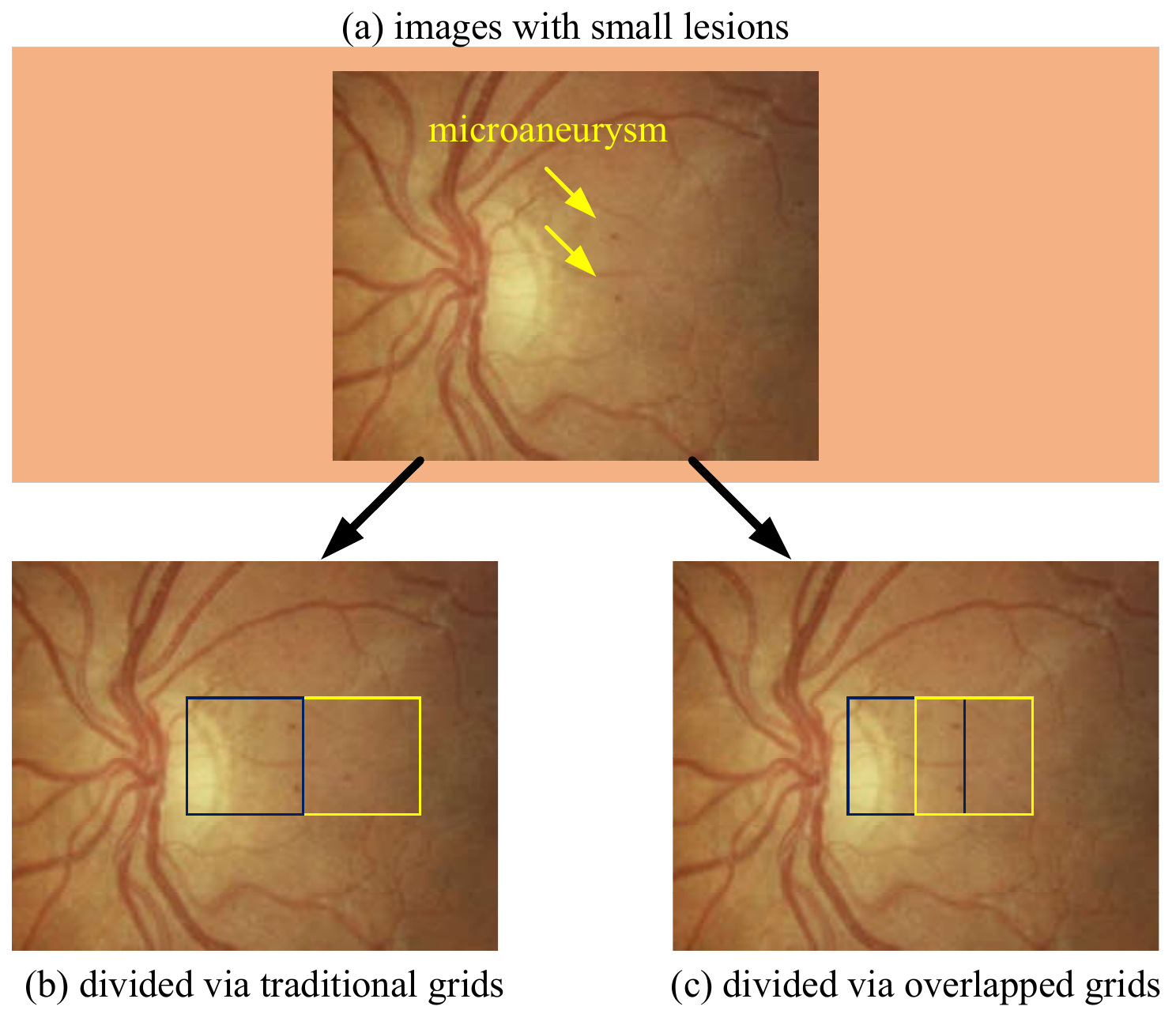}
\caption{The advantages of overlapping grids. Compare to (b), The yellow grid in (c) can alleviate the drawbacks of traditional non-overlap grids when small lesions are located in the edge of grid.}
\label{fig.overlapGrid}       
\end{figure}

A straightforward method to keep the relative locations is partitioning via grids. However, as shown in Fig. \ref{fig.overlapGrid}, when microaneurysm or some small hemorrhage happen to located in the edge of the grid, they may be misclassified into normal patches. In this paper, we solve this issue by dividing the images with overlapped grids, which is a special case of sliding windows. The grid size is fixed as $h \times h$ (in pixels), and the sliding stride is $h-ov$, where $ov$ is the overlapped size between two adjacent grids. As seen in Fig. \ref{fig.overlapGrid}(c) the small lesions are located at more conspicuous place in the yellow grid than those in Fig. \ref{fig.overlapGrid}(b). Therefore, the proposed dividing method can retain the relative location of the image patches, as well as generating high quality candidates for lesions detection.

\subsection{Local Network} \label{subsec.lessionMap}

In the last subsection, the input images are divided into $h \times h$ patches. Our local network is trained to classify the patches into $0$ (normal), $1$ (microaneurysm), $2$ (hemorrhage), $3$ (exudate), which are the main indicators to NPDR.


Inspired by the outstanding works in deep convolutional neural network \cite{Alexnet,Googlenet,ResNet}, the proposed network is composed by convolutional layer, max-pooling layer, fully connected (FC) layer. The activation function between two layers is rectified linear unit (ReLU), and batch normalization (BN) is applied before each ReLU. Additionally, Dropout \cite{Dropout} is utilized after FC layers. The label of the input patch is given by the output of soft-max regression \cite{PRML}.  The detailed architecture of the local network is presented in TABLE \ref{tab.localNetwork}.


\begin{table}[!htb]
\centering
\caption {Local Network Architecture}
\setlength{\tabcolsep}{5.5pt}
\begin{tabular}{cccc}
\hline
Layer  & Type & kernel size and number & stride \\
\hline
0 & input & ... & ... \\
\hline
1 & convolution & 3 $\times$ 3 $\times$ 64 & 1 \\
\hline
2 & convolution & 3 $\times$ 3 $\times$ 128 & 1 \\
\hline
3 & max-pooling & 2 $\times$ 2 & 2 \\
\hline
4 & convolution & 3 $\times$ 3 $\times$ 128 & 1 \\
\hline
5 & max-pooling & 2 $\times$ 2 & 2 \\
\hline
6 & convolution & 3 $\times$ 3 $\times$ 256 & 1 \\
\hline
7 & fully connected & 1 $\times$ 1 $\times$ 512 & ... \\
\hline
8 & fully connected & 1 $\times$ 1 $\times$ 1024 & ... \\
\hline
9 & soft-max & ... & ... \\
\hline
\label{tab.localNetwork}
  \end{tabular}
\end{table}

\subsection{Weighed Lesion Map}

\begin{figure}[!t]
\centering
\includegraphics[width=3in]{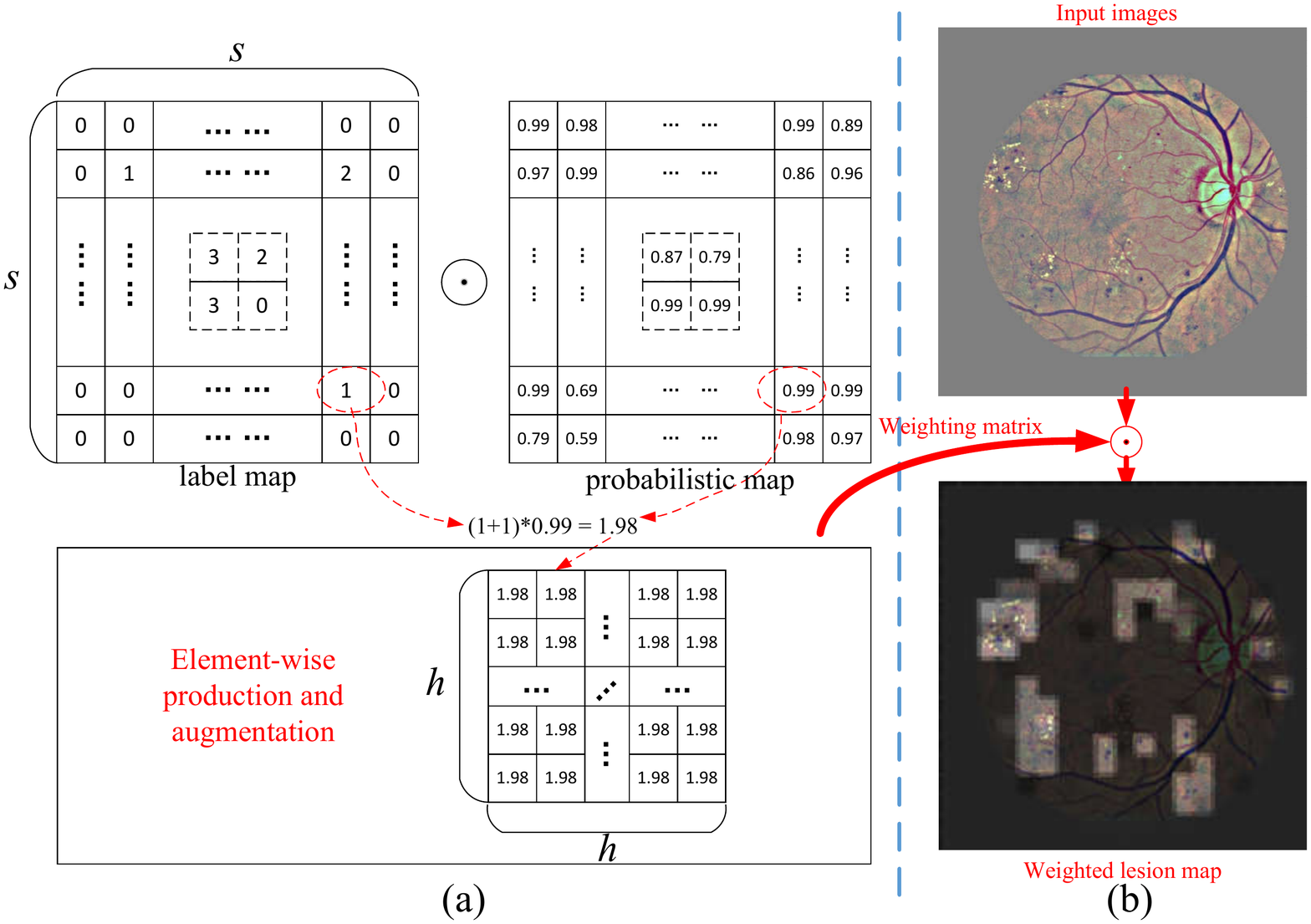}
\caption{The illustration of the construction of weighted lesion map. (a) Jointing label map and probabilistic map into weighting matrix. (b) Illustration of imbalanced attention on weighted lesion map.}
\label{fig.lesionMap}       
\end{figure}

Two maps are generated when all the patches in fundus image ${\bf I} \in \mathbb{R} ^ {d \times d}$ are classified by the local network. One is label map ${\bf L} \in \mathbb{R} ^ {s \times s}$, which records the predicted labels of the patches. Wherein $s  = \lfloor (d - h)/(h-ov) \rfloor$, and $\lfloor . \rfloor$ is the floor operator. The other map is probabilistic map ${\bf P} \in \mathbb{R} ^ {s \times s}$, which retains the the biggest output probability of the softmax layer for each patch label. Based on these two maps, we construct a weighting matrix for each input image as follows: (1) Integrating the label map and probabilistic map as $ {\bf LP} = ({\bf L+ 1}) \odot {\bf P}$, where $\odot$ is the element-wise product and ${\bf 1} \in \mathbb{R} ^ {s \times s}$ is an all one matrix \footnote{The motivation of the addition of the all one matrix is to avoid totally removing the information in the patches with label $0$.}. (2) Each entry in $\bf{LP}$ is augmented to a $h \times h$ matrix (corresponding to the patch size), and the illustration of (1) and (2) step can be seen in Fig. \ref{fig.lesionMap}(a). (3) Tiling the augmented matrixes into the weighting matrix ${\bf M_I} \in \mathbb{R} ^ {d \times d}$ according to the relative locations in the input image ${\bf I}$. The weighting matrix is constructed into the same size of input image, and the values in the intersection areas are set as the average values between adjacent expanded matrixes.

The weighted lesion map of input image ${\bf I}$  is defined as ${\bf I}^* = {\bf M_I} \odot {\bf I}$. The entries in the weighting matrix ${\bf M_I}$ implicit the severity and probability of lesions in local patches. Therefore, the image patches have more severe lesion patches with higher probability will get higher weights in the weighted lesion map. As seen in Fig. \ref{fig.lesionMap}(b), imbalanced attentions are payed on the weighted lesion map by highlighting the lesion patches.

By feeding the weighted lesion map into the global network to grade the DR severity of the fundus images, the proposed grading network trends to pay more attention on the patches with severer lesions. Therefore, as shown in the experimental section, the grading algorithms with the proposed weighting scheme outperform the traditional end-to-end grading networks under the same implementation setup.

\subsection{Global Network} \label{subsec.globalNetwork}

The global network is designed to grade the severity of DR according to the {\em International
Clinical Diabetic Retinopathy scale} \cite{DRstages}. Since PDR is the most severe stage in the standard, it's somewhat too late to precaution and remedy when the patients already step into PDR. Therefore, in this study, we focus on the advanced detection on NPDR. The rough description of the stages of fundus images are characterized as:

\begin{itemize}
\item class 0 - No visible lesions and abnormalities.
\item class 1 - Mild NPDR, only microaneurysms.
\item class 2 - Moderate NPDR, extensive microaneurysms, haemorrhages, and hard exudates.
\item class 3 - Severe NPDR, venous abnormalities, large blot haemorrhages, cotton wool spots, venous beading, venous loop, venous reduplication
\end{itemize}

The global network is trained with weighted lesion map, and the output is the severity grade of the testing fundus images. Since the size of weighted lesion map is bigger than the patches for local network, the depth of the global network is deeper than the local one. The detailed architecture of the global network is shown in TABLE \ref{tab.globalNetwork}. Similar with local network, BN is used before each ReLU activation, and Dropout is implemented after fully connection layers.

\begin{table}[!htb]
\centering
\caption {Global Network Architecture}
\setlength{\tabcolsep}{5.5pt}
\begin{tabular}{cccc}
\hline
Layer  & Type & kernel size and number & stride \\
\hline
0 & input & ... & ... \\
\hline
1 & convolution & 3 $\times$ 3 $\times$ 32 & 1 \\
\hline
2 & max-pooling & 2 $\times$ 2 & 2 \\
\hline
3 & convolution & 3 $\times$ 3 $\times$ 32 & 1 \\
\hline
4 & max-pooling & 2 $\times$ 2 & 2 \\
\hline
5 & convolution & 3 $\times$ 3 $\times$ 64 & 1 \\
\hline
6 & max-pooling & 2 $\times$ 2 & 2 \\
\hline
7 & convolution & 3 $\times$ 3 $\times$ 64 & 1 \\
\hline
8 & max-pooling & 2 $\times$ 2 & 2 \\
\hline
9 & convolution & 3 $\times$ 3 $\times$ 128 & 1 \\
\hline
10 & max-pooling & 2 $\times$ 2 & 2 \\
\hline
11 & convolution & 3 $\times$ 3 $\times$ 128 & 1 \\
\hline
12 & max-pooling & 2 $\times$ 2 & 2 \\
\hline
13 & convolution & 3 $\times$ 3 $\times$ 256 & 1 \\
\hline
14 & max-pooling & 2 $\times$ 2 & 2 \\
\hline
15 & convolution & 3 $\times$ 3 $\times$ 256 & 1 \\
\hline
16 & max-pooling & 2 $\times$ 2 & 2 \\
\hline
17 & convolution & 3 $\times$ 3 $\times$ 512 & 1 \\
\hline
18 & max-pooling & 2 $\times$ 2 & 2 \\
\hline
19 & convolution & 3 $\times$ 3 $\times$ 512 & 1 \\
\hline
20 & fully connected & 1 $\times$ 1 $\times$ 1024 & ... \\
\hline
21 & fully connected & 1 $\times$ 1 $\times$ 1024 & ... \\
\hline
22 & fully connected & 1 $\times$ 1 $\times$ 4 & ... \\
\hline
23 & soft-max & ... & ... \\
\hline
\label{tab.globalNetwork}
  \end{tabular}
\end{table}

\section{Experiments, Results and Discussions}

\subsection{Datasets}

{\em Kaggle database}: The Kaggle database contains $35,126$ training fundus images and $53,576$ testing fundus images. All the images are assigned into five DR stages, i.e., four NPDR stages as presented in Section \ref{subsec.globalNetwork} and one stage denotes PDR. The images in the dataset come from different models and types of cameras under various illumination.

According to our cooperate ophthalmologists, although the ammont of images in this dataset is relatively big, there exist a large portion of biased labels. Additionally, the dataset do not indicates the locations of the lesions which are meaningful to clinicians. Therefore, we select subset from Kaggle database for re-annotation. The subset consists of $22,795$ randomly selected images in terms of the four grades of NPDR, where $21,995$ for training and $800$ for testing (each NPDR grade contains $200$ testing images).

The training and testing patches for lesion detection are cropped from the training and testing images respectively, which totally contains $12,206$ lesion patches and over 140 thousands randomly cropped normal patches. Licensed ophthalmologists and trained graduate students are invited or payed to annotate the lesions in the images and re-annotate DR grades of the fundus images. The detailed statistics of the DR grades and patch labels can be seen in TABLE \ref{tab.kaggleSta} and \ref{tab.kagglePatchSta}.



\subsection{Data Preparation}

\begin{figure}[!t]
\centering
\includegraphics[width=3in]{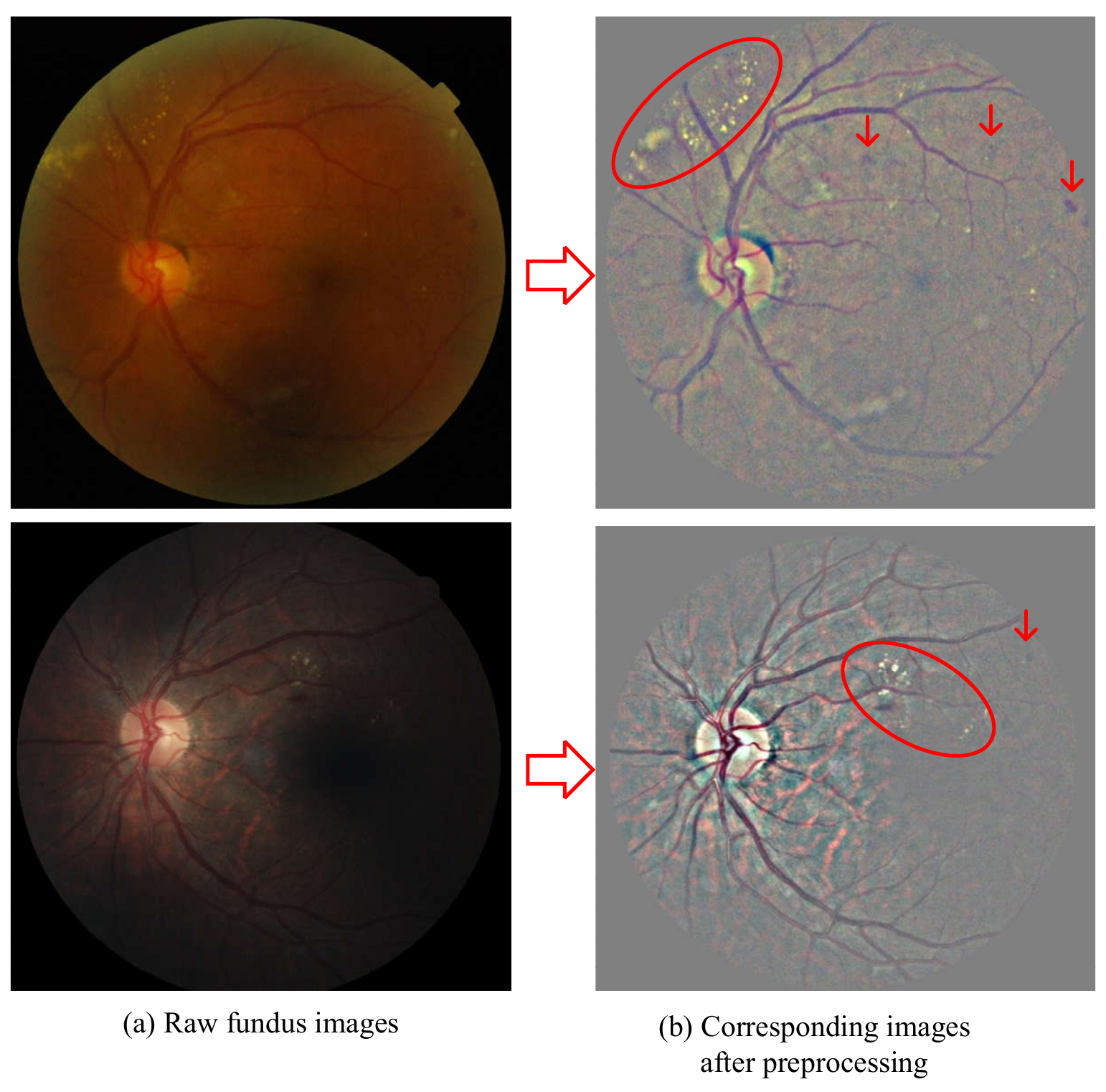}
\caption{Comparison between the raw fundus images before and after preprocessing. The red ellipse and arrows in the right indicate some underlying lesions can be singularized after preprocessing.}
\label{fig.preprocess}       
\end{figure}

1) {\em Data Preprocessing.} Inspired by \cite{TMI2016}, contrast improvement and circular region of interesting extraction are conducted on the color fundus images as following:

\begin{equation}
{\bf I} = \alpha {\bf I}_{raw} + \beta {\bf G}(\theta) \circ {\bf I}_{raw} + \gamma,
\label{eq.preprocess}
\end{equation}
where  ${\bf I}_{raw}$ is the raw fundus image in the datasets, and ${\bf I}$ denotes the corresponding output image after preprocessing. ${\bf G}(x,y;\theta)$ is a Gaussian kernel with scale $\theta$. In this paper, $\theta$ is empirically set as $10$. $\circ$ is the convolution operator. $\alpha$, $\beta$ and $\gamma$ are parameters to control the weights of different parts in Eq. \eqref{eq.preprocess}, which are respectively set as $4$, $-4$ and $128$. As seen in Fig. \ref{fig.preprocess}, compared to the raw fundus images, some underlying lesions can be easily exploited after the preprocessing step.

Additionally, for lesion detection, all the images are resized to $800 \times 800$, and the relative ratio between the sample height and length is kept by padding before resizing the raw images. The input sample size of global network are turned into $256 \times 256$ to reduce the computational complexity.

2) {\em Data Augmentation.}

Data augmentation are implemented to enlarge the training samples for deep learning, as well as to balance the samples across different classes. Inspired by \cite{Alexnet}, the augmentation methods include randomly rotation, cropping and scaling. The detailed statistics of the dataset are presented in TABLE \ref{tab.kaggleSta} and TABLE \ref{tab.kagglePatchSta}. The images after augmentation is split into training and validating set for tuning the deep models, and the testing images presented in the last line of TABLE \ref{tab.kaggleSta} are not put into augmentation.

\begin{table}[!htb]
\centering
\caption {Statistics of fundus images from Kaggle dataset}
\setlength{\tabcolsep}{5.5pt}
\begin{tabular}{lcccc}
\hline
     & Normal  & Mild & Moderate  & Severe  \\
\hline
Raw	images & 18472 &	1870 &	1870&	583\\
Augmentation	&18472&	28050	&29869&	27401\\
Total number	&36944	&29920&	34136&	27984\\
\hline
Train number&	35844&	28920&	33136&	26984\\
Validation number&	1000&	1000&	1000&	1000\\
\hline
Test number & 200 & 200 & 200 & 200  \\
\hline
\label{tab.kaggleSta}
  \end{tabular}
\end{table}

\begin{table}[!htb]
\centering
\caption {Statistics of lesion patches from Kaggle dataset}
\setlength{\tabcolsep}{5.5pt}
\begin{tabular}{lcccc}
\hline
     & Normal  & MA & Hemorrhage  & Exudate  \\
\hline
Raw	patches & 1373&	1945&	2144	&1614\\
Augmentation	&12147&	8275	&10043&	9823\\
Total number	&13520&	10220&	12187&	11437\\
\hline
Train number&	13220&	9920&	11887&	11137\\
Validation number&	300	&300&	300&	300\\
\hline
\label{tab.kagglePatchSta}
  \end{tabular}
\end{table}

3) {\em Reference Standard and Annotation.} In this paper, the reference of the samples are generated from the integration of three type of observers: the original annotations from Kaggle competition, the trained graduate students \footnote{All the trained observers have experience in medical image processing} and a clinical ophthalmologist.

Concretely, for training the local network, the patches (TABLE \ref{tab.kagglePatchSta}) are first annotated by two over three months trained observers, then all the samples are checked by clinical ophthalmologists. For training the global network, the labels of the fundus images are selected from Kaggle annotations by the trained observers, then the label biases are corrected by the clinical ophthalmologist.

For the testing sets, firstly, all the testing lesion patches and DR grades of fundus images are annotated independently among all the trained observers and ophthalmologist, then the discrepancy patches are selected for further analyzing. Finally, the references of the patches are determined only by achieving the agreement of all annotators.

\subsection{Evaluation design and Metrics}

To clearly evaluate the performance of the proposed algorithm, we conduct the following experiments:
(1) For analyzing the effectiveness of the local network, we record the recall and precision of each type of lesions in testing images. Additionally, the sensitivity and specificity between normal patches and lesion patches are shown with receiver operating characteristics (ROC) curve and area under curve (AUC) metric. (2) To show the superior of the proposed algorithm in grading the severity of fundus images,  {\em Kappa score} \footnote{{\em Kappa score} typically varies from $0$ (random agreement between raters) to $1$ (complete agreement between raters)} \cite{IRbook} and classification accuracy are introduced to measure the agreement between the predictions and the reference grades. In this paper, the calculation of {\em Kappa} metric is same with Kaggle Diabetic Retinopathy Detection competition Readers can see the following link  for more details: {\em https://www.kaggle.com/c/diabetic-retinopathy-detection/details/evaluation}. Besides, ROC curve and AUC are also took into comparison between different grading algorithms to classify normal fundus images and referable DR (moderate or worse DR) images.

To further clarify the evaluation metrics, we list some necessary conceptions as following.

\noindent For multi-class classification:
\begin{itemize}
\item {\em Recall} - For class $i$, recall means the rate of correct predicted number over total number of class $i$ in the testing set.
\item {\em Precision} - For class $i$, precision means the rate of correct predicted number over total number of class $i$ predicted by the algorithm.
\item {\em Accuracy} - The total number of correct predictions divide the number of samples in testing set.
\end{itemize}
\noindent For binary classification between lesion (positive) and normal (negative) patches, as well as between referable DR (positive) and normal (negative) fundus images.
\begin{itemize}
\item {\em True positive (TP)} - The number of positive sample is correctly detected.
\item {\em False positive (FP)} - The number of negative sample is miss classified as positive.
\item {\em False negative (FN)} - The number of positive samples is miss classified as negative.
\item {\em True negative (TN)} - The number of negative sample is correctly detected.
\end{itemize}

Based on the aforementioned description, {\em sensitivity} and {\em specificity} is defined as
\begin{equation}
\begin{array}{l}
sensitivity = \frac{TP}{TP+FN}, \\
\\
specificity = \frac{TN}{FP+TN}.
\end{array}
\end{equation}

\subsection{The Identification of Lesions}

To evaluate the performance of local network for lesion recognition, we record the recall and precision for each class of lesion in testing fundus images, and the statistics of lesions in testing images can be found in TABLE \ref{tab.lesion_RP}. The second line of the table present the number of different types of lesions in testing set. The left and right values in the table denote {\em recall} and {\em precision} respectively. Two baseline algorithms are take into comparison: random forests (RF) \cite{randomForest} and support vector machine (SVM) \cite{SVM}. The input of these two alternative algorithms are image patches with are in the same condition with our local network, and we use the default setting with a {\em Python} toolkit named {\em Sciket-learn} \footnote{http://scikit-learn.org/stable/.} except that the number of RF trees is turned from 10 to 500 (Readers can see the url in the footnote for Details). As seen in the table, the proposed local network significantly outperforms the random forests and SVM under same training images, which indicate the powerful ability of DCNN in learning task-driven features.

\begin{figure}[!t]
\centering
\includegraphics[width=3in]{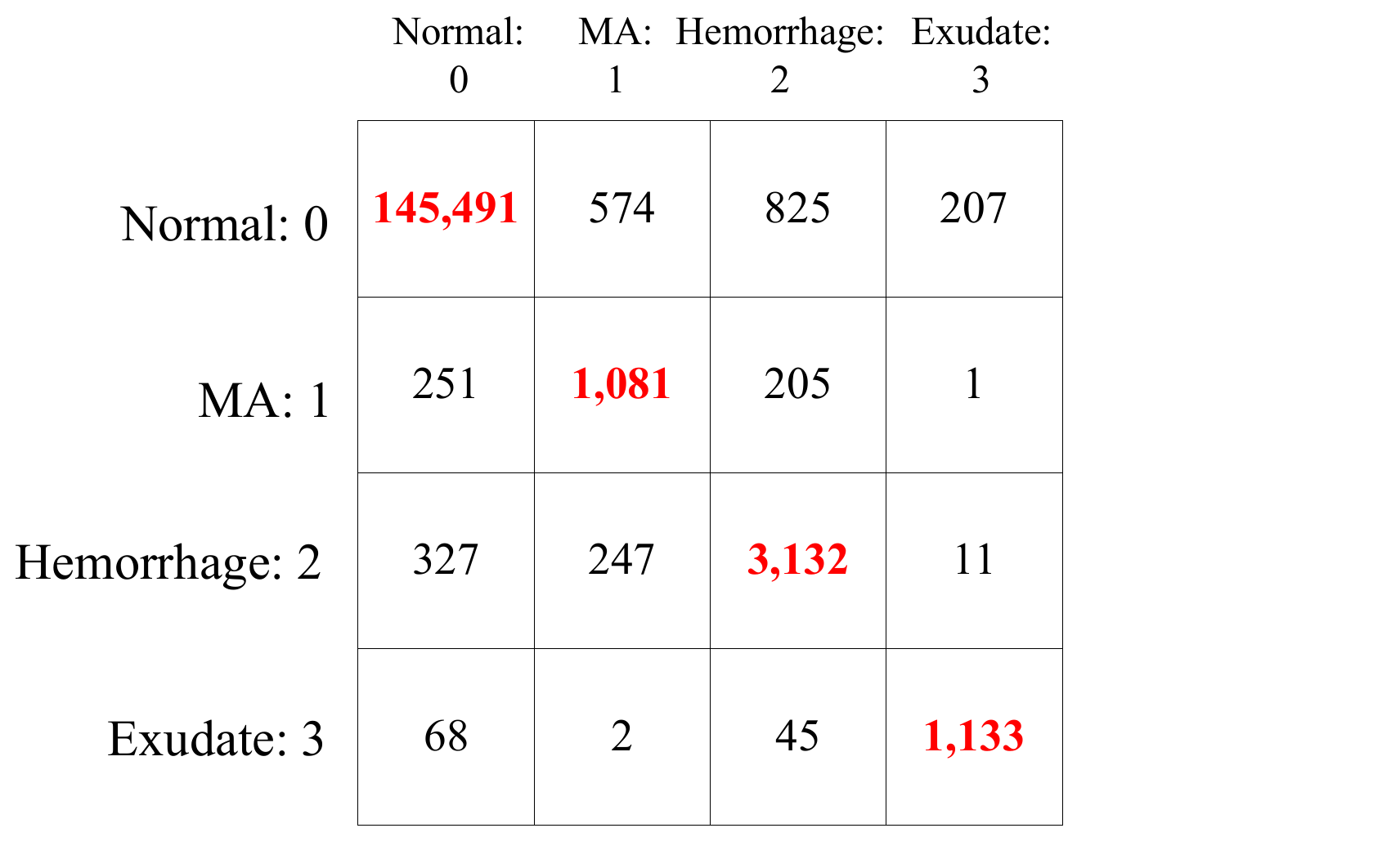}
\caption{Lesion confusion matrix. The value of $(i,j)$-th entry of the matrix denotes the number of class $i$ patches with prediction as class $j$. Wherein, $i,j \in \{0,1,2,3\}$ according to the first and second axes respectively. The number of correct prediction with respect to each type of lesion is shown in red.}
\label{fig.lesionConfusionMatrix}       
\end{figure}

As listed in TABLE \ref{tab.lesion_RP}, the {\em recall} and {\em precision} of MA and hemorrhages are relatively worse than exudate in the proposed algorithm. The reason behind may be three folders: (1) the MA is too inconspicuous and small for detecting even with our own eyes. (2) The MAs and some small hemorrhages are such similar that it is challenging to distinguish them even if the lesions are detected. (3) Compared to testing set, the annotations of training set are less strict. Therefore, the labels between some MAs and hemorrhages may mixed up in the training of local network, which further effect the recognition between these two types of lesions. As seen the confusion matrix for lesion recognition in Fig. \ref{fig.lesionConfusionMatrix}, there are lots of MA and hemorrhages are successfully detected but mis-classified. Additionally, some cotton wool spots are detected as exudates in the experiments, which lead to some false positives in exudates detection. This issue can be addressed by adding cotton wool spots into our training set in the further work.

\begin{table}[!htb]
\centering
\renewcommand{\arraystretch}{1.1}
\caption {Recall (the left values) and precision (the right values) of lesion recognition}
\setlength{\tabcolsep}{5.5pt}
\begin{tabular}{lccc}
\hline
  & MA & Hemorrhage  & Exudate  \\
\hline
lesion number & 1538 & 3717 & 1248  \\
\hline
local network & \textbf{0.7029 / 0.5678}  & \textbf{0.8426 / 0.7445} & \textbf{0.9079 / 0.8380} \\

Random Forest & 0.0078 / 0.06704 & 0.2754 / 0.1011 & 0.6859 / 0.1941 \\

SVM & 0.4153 / 0.0251 & 0.0108 / 0.0548 &  0.0787 / 0.0318 \\
\hline
\label{tab.lesion_RP}
  \end{tabular}
\end{table}

\begin{figure}[!t]
\centering
\includegraphics[width=3in]{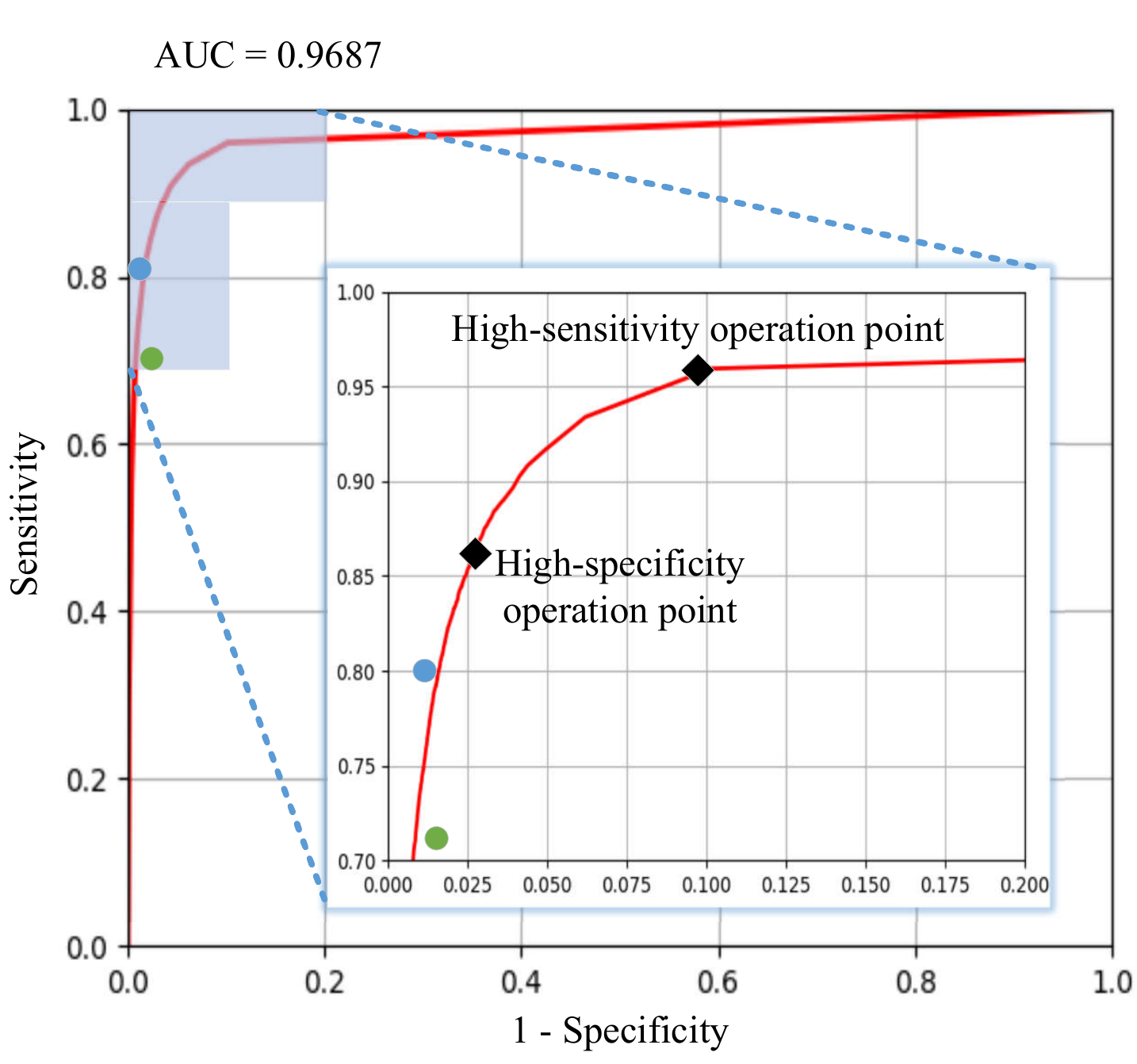}
\caption{ROC curve (shown in red) of the proposed algorithm over lesion detection. The black diamonds on the curve indicate the sensitivity and specificity of our lesion detection algorithm on high-sensitivity and high-specificity operating points. The green and blue dots present the performance of two trained human observers on binary lesion detection on the same testing dataset.}
\label{fig.patchROC}       
\end{figure}

To show the importance of local net in detecting the lesions, we also train a binary classifier to distinguish the lesion patches from normal ones in the testing set. ROC curve is drawn with {\em sensitivity} and {\em specificity} in Fig. \ref{fig.patchROC}, and the value of AUC is $0.9687$. The black diamonds on the red curve highlight the performance of the proposed algorithm at high-specificity ($sensitivity: 0.863, specificity: 0.973$) and high-sensitivity point ($sensitivity: 0.959, specificity: 0.898$). The green and blue dots correspond to  the $sensitivity$ and $1-specificity$ of two trained observers on binary lesion detection. As shown in the figure, the proposed algorithm can achieve superior performance over the trained human observers by setting a proper operating point.

\subsection{Grading The Severity of Fundus Images} \label{sec.exp_dr_grading}

\begin{figure}[!t]
\centering
\includegraphics[width=3in]{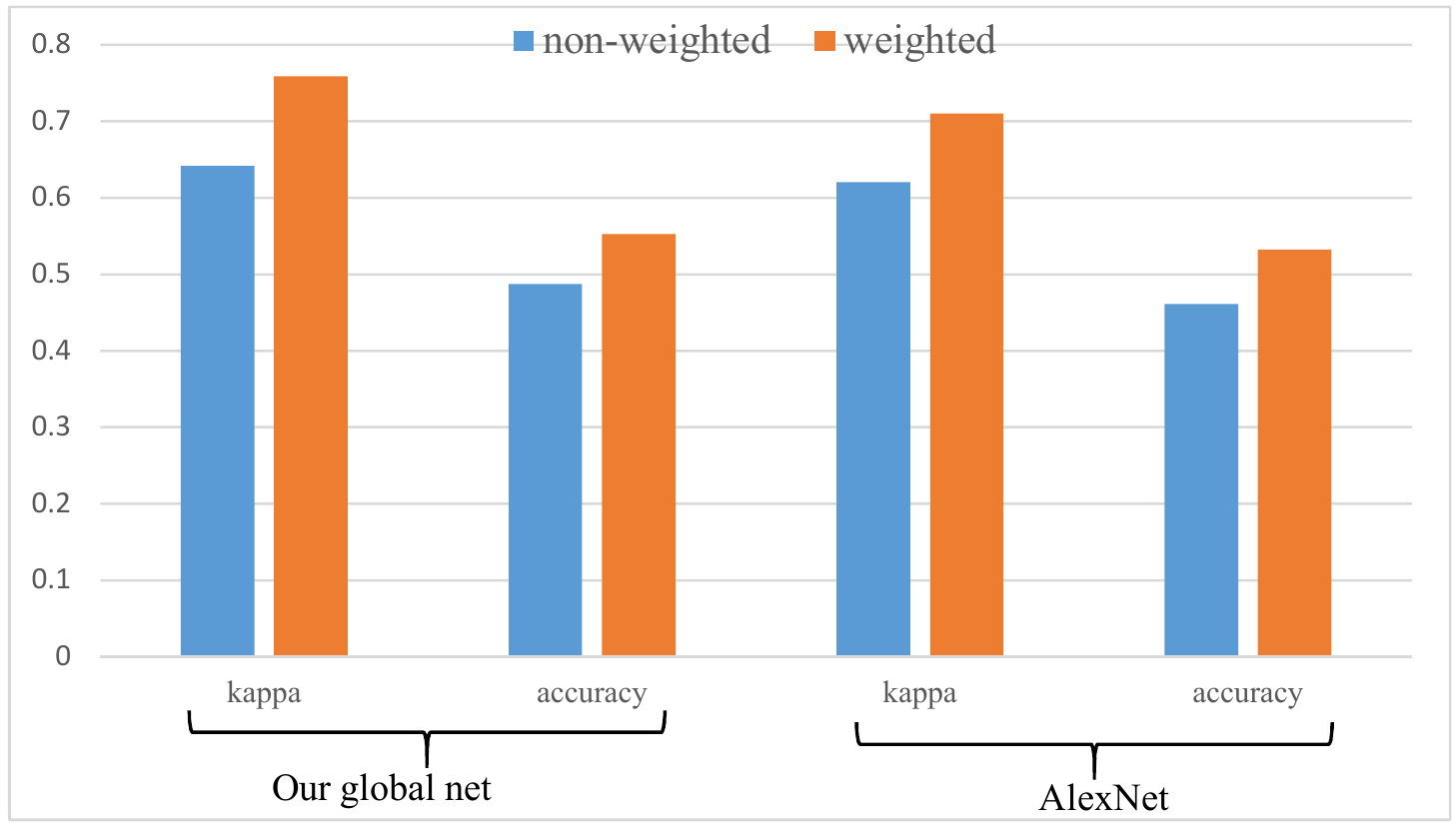}
\caption{Illustration of importance of the proposed grading net. As seen in the figure, both grading nets achieve performance enhancement in kappa and accuracy scores with the proposed weighted scheme.}
\label{fig.grading_kappa_acc}       
\end{figure}

In this paper, we focus on the grading on NPDR, which can be classified into $0$ to $3$ stages: normal, mild, moderate and severe respectively. To prove the importance of the proposed weighting scheme, we compare the {\em Kappa score} and {\em Accuracy} of grading networks with and without weighting ({\em non-weighted} for simplification) scheme under the same implementation setup. The results are shown in Fig. \ref{fig.grading_kappa_acc}. As seen in the figure, both our global net and the popular AlexNet achieve superior results with weighted lesion map, which prove the effectiveness of the proposed weighted scheme.

\begin{figure}[!t]
\centering
\includegraphics[width=3in]{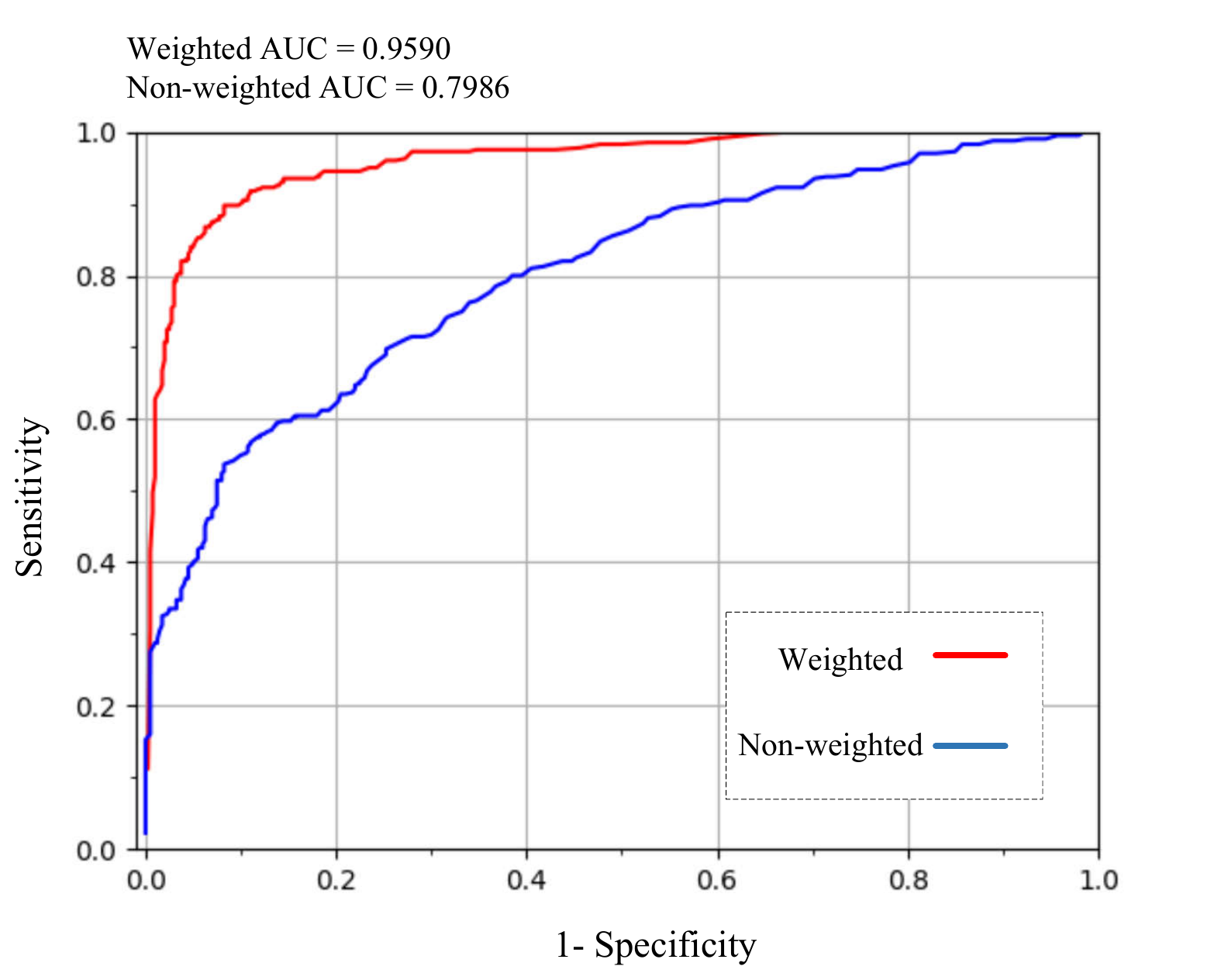}
\caption{ROC curves of grading network in referable DR detection under weighted (red) and non-weighted (blue) conditions. }
\label{fig.grade_roc}       
\end{figure}

Since the symptom of some milder DR are too unconspicuous to be spotted, the judgements of milder DR are not easy to be unified even among licensed ophthalmologists. Therefore, similar with \cite{googleJAMA}, we also train our global net to distinguish referable DR from normal images. The ROC curves on {\em sensitivity} and  {\em 1-specificity} are illustrated in Fig. \ref{fig.grade_roc}. The performance of referable DR detection with weighted scheme is shown in red, and the AUC of the proposed algorithm is $0.9590$. On the other side, the performance of the same network under non-weighted scheme is shown in blue, and the corresponding AUC is $0.7986$. The comparison results further prove the effectiveness of the proposed weighted scheme in DR grading.

%

\section{Discussion}

The experimental results in the last section prove the effectiveness of the proposed local network in lesion detection and recognition, and the outperformance of the global net with weighting lesion map indicate the superiority of the proposed imbalanced weighting scheme for DR grading.
On the one hand, the lesion information from weighting matrix promote the performance of the proposed global grading net. On the other hand, by learning the underlying feature in both images and weighting matrix, the global deep neural network smooth some mistakes generated by the local net in grading the severity of fundus images. To sum up, both the global and local network are two complementary parts to provide more abundant information for ophthalmologists and patients in DR analysis.

Although the proposed study gives reasonable results in lesion detection and NPDR grading, there still exists some imperfections, such as the quantity and quality of the carefully annotated training samples need to be improved.
It will be better if more professional observers can plug into sample annotation.
However, limited by the time and current resources, we've tried our best to ensure the correctness of the samples and show the performance of the algorithm reasonably. In addition, we only detect lesions including MA, hemorrhage and exudate, but some other important abnormal observations are not addressed (e.g., venous beading in NPDR and neovascularization in PDR). However, benefited from the powerful ability of deep learning, the features of these untreated lesions may be learned for grading the severity of DR \cite{googleJAMA}.

In addition, we aim to provide a two-stages networks pipeline and imbalanced weighting framework for DR analysis in this paper. Therefore, the global and local networks can be any architecture which can achieve better performance.




\section{Conclusion}

In this paper, we proposed two-stages DCNN to detect abnormal lesions and severity grades of DR in fundus images. The experimental results have shown the effectiveness of the proposed algorithm, and this study can provide valuable information for clinical ophthalmologists in DR examination. However, there still exist limitations need to be solved in our future work, such as collecting more high quality annotated fundus data, and paying attention to more types of lesions. Moreover, diabetic macular edema is also an import open issue needed to be addressed.



%


\section*{Acknowledgment}

First, we would like to thank {\em Kaggle Diabetic Retinopathy Detection} competition for providing the valuable fundus images datasets to promote our research. Second, many thanks for some anonymous ophthalmologists, annotators and researchers who give helpful comments and advices for this paper.

\ifCLASSOPTIONcaptionsoff
  \newpage
\fi

\end{document}